\documentclass{article}
\usepackage{spconf,amsmath,graphicx}
\usepackage{amssymb,bm,booktabs,cite}
\usepackage[hyphens]{url}
\usepackage{balance}

\usepackage{enumitem}
\setlist{nosep, leftmargin=14pt}

\def\L{{\cal L}}
\newcommand{\BCE}{\operatorname{BCE}}
\newcommand{\FullMicro}{0.338}

\newcommand{\StaticMicro}{0.302}

\newcommand{\GainStatic}{0.036}
\newcommand{\GainStaticLo}{0.022}
\newcommand{\GainStaticHi}{0.049}

\newcommand{\GainStaticSeedMin}{0.019}
\newcommand{\GainStaticSeedMax}{0.057}
\newcommand{\StaticOverManual}{0.017}
\newcommand{\StaticManualLo}{0.004}
\newcommand{\StaticManualHi}{0.030}

\newcommand{\CtrlNearStatic}{0.006}
\newcommand{\FullOverCtrlMin}{0.034}

\newcommand{\LagOverStatic}{0.018}
\newcommand{\LagGap}{0.023}
\newcommand{\LagGapLo}{0.004}
\newcommand{\LagGapHi}{0.040}
\newcommand{\CacheHitFull}{42}
\newcommand{\NSeedsMain}{5}

\newcommand{\SubsetFullGain}{0.020}
\newcommand{\SubsetManualGain}{0.027}

\newcommand{\SubsetCfourVsFull}{+0.007}
\newcommand{\SubsetCfourVsFullLo}{$-$0.009}
\newcommand{\SubsetCfourVsFullHi}{+0.024}

\newcommand{\AUCRangeLo}{0.809}
\newcommand{\AUCRangeHi}{0.815}

\newcommand{\UngroundedFull}{6.1}
\newcommand{\UngroundedStatic}{7.4}
\newcommand{\GPUhFull}{34.4}

\newcommand{\StudentParams}{11.1}
\newcommand{\StudentGMACs}{5.7}
\newcommand{\StudentLatency}{9.6}

\newcommand{\GemmaSpeedRatio}{200}
\newcommand{\TeacherCostRatio}{5.8}
\newcommand{\PopStudies}{915}
\newcommand{\PopPos}{341}
\newcommand{\PopBoxes}{402}
\newcommand{\PopNeg}{1{,}691}
\newcommand{\DistillImages}{40{,}000}
\newcommand{\Refreshes}{10}

\newcommand{\PilotBestAUROC}{0.781}

\newcommand{\PilotOracle}{0.71}
\newcommand{\PilotTopOne}{0.55}
\newcommand{\RuleCoverage}{58}
\newcommand{\GenValid}{88.9}

\newcommand{\MovedDrop}{0.24}
\newcommand{\CommonThrGain}{0.035}
\newcommand{\CommonThrLo}{0.021}
\newcommand{\CommonThrHi}{0.049}
\newcommand{\DualReaderGain}{0.037}
\newcommand{\DualReaderLo}{0.020}
\newcommand{\DualReaderHi}{0.055}

\newcommand{\EDMPairs}{12}
\newcommand{\PTXPairs}{10}
\newcommand{\NihBoxes}{577}
\newcommand{\NihImages}{550}
\newcommand{\NihPatients}{471}

\newcommand{\NihStatic}{0.423}
\newcommand{\NihFull}{0.447}

\newcommand{\NihStaticGain}{0.007}
\newcommand{\NihStaticGainLo}{$-$0.005}
\newcommand{\NihStaticGainHi}{0.019}
\newcommand{\NihAdaptGain}{0.024}
\newcommand{\NihAdaptGainLo}{0.011}
\newcommand{\NihAdaptGainHi}{0.038}

\title{APED-CXR: Adaptive Privileged Evidence Distillation for Chest X-Ray Grounding with Image-Only Inference}
\name{Md Shihabul Islam Shovo$^{\star}$ \qquad Nishi Kanta Paul$^{\star}$ \qquad Jamil Fayyad$^{\dagger}$}
\address{$^{\star}$NIMISHES Lab, Dhaka, Bangladesh \qquad
         $^{\dagger}$American University of Sharjah, Sharjah, UAE \\
         {\small shihabul900@gmail.com, nishikantapaul108@gmail.com, jfayyad@aus.edu}}
\begin{document}
\ninept
\maketitle
\begin{abstract}
Compact chest X-ray (CXR) models that localize findings suit deployment, but box annotations are limited and report-aware multimodal models are expensive at inference. We introduce APED-CXR, which distills report-informed spatial evidence from a frozen multimodal teacher into a student with \StudentParams M parameters that predicts diseases and boxes from shared evidence maps and requires the image alone at test time. During training, the teacher uses the radiograph, report, and queried class to evaluate regions proposed from the student's current evidence maps, and its support scores serve as training targets for the student's pooled evidence scores within these regions. Trained on PadChest and evaluated on held-out PadChest-GR, refreshing the student's proposals for teacher evaluation improves box-level localization F1 from \StaticMicro{} with fixed warm-up proposals to \FullMicro{} (IoU\,$\geq$\,0.5; gain \GainStatic, 95\% CI [\GainStaticLo, \GainStaticHi]), while maintaining diagnostic AUC. This gain depends on the teacher checking current proposals: with equally many random boxes, a fixed dense box set, teacher-generated boxes, or report rules instead of the teacher, F1 stays near the fixed-proposal level, and proposals two refreshes old recover only part of the gain. On NIH ChestX-ray14, a secondary test set, box recall also rises from \NihStatic{} to \NihFull.
\end{abstract}


\section{Introduction and Related Work}
\label{sec:intro}

Automated chest X-ray (CXR) interpretation is increasingly expected to localize where a finding occurs, not only whether it is present. For routine use, such a model should also be compact and independent of the radiology report, which does not yet exist when the image is first read. Report-aware vision--language models (VLMs) can localize findings, but are too large and text-dependent for use at the time of the read, while compact image-only models rarely receive enough box supervision to do so reliably. We thus introduce \textbf{A}daptive \textbf{P}rivileged \textbf{E}vidence \textbf{D}istillation for \textbf{C}hest \textbf{X}-\textbf{R}ays (APED-CXR), in which a frozen multimodal teacher verifies the student's proposed regions during training only, distilling report-informed spatial knowledge into a compact image-only model.

Prior CXR localization techniques mostly used image-level supervision. Class activation maps (CAM)~\cite{zhou2016cam} project classifier weights onto convolutional features to highlight class-discriminative regions, while pooling-based multiple-instance learning~\cite{pinheiro2015image,wang2017chestxray8} combines local predictions into image-level scores, for example with log-sum-exp pooling; ChestX-ray8~\cite{wang2017chestxray8} applied the approach to text-mined labels and evaluated localization on a small boxed subset. However, a saliency benchmark found such maps inferior to radiologists at localizing pathology~\cite{saporta2022benchmarking}, which motivates supervising the evidence map directly. Li et al.~\cite{li2018thoracic} showed that a few boxes can supplement image-level supervision, but expert-curated datasets are small: PadChest-GR provides radiologist boxes in 4{,}555 CXR studies~\cite{castro2025padchestgr}, while PadChest provides over 160{,}000~\cite{bustos2020padchest}. Thus, most training samples have labels and reports but lack boxes.

Recent VLMs have used reports to localize findings. BioViL~\cite{boecking2022making} grounds phrases using image--report pairs, CoNNS~\cite{lian2026conns} suppresses noisy negatives for zero-shot classification and grounding, while MAIRA-2~\cite{bannur2024maira2} and ChEX~\cite{muller2024chex} generate grounded reports or region descriptions. MedGemma~\cite{sellergren2025medgemma} answers localized questions about radiographs. Although these advances exist, such models require text at inference or have billions of parameters, which limits their use as compact first-read tools. RegionCLIP~\cite{zhong2022regionclip} suggests another route: a vision--language model assigns scores to region proposals, creating pseudo-labels so that its knowledge supervises a separate model instead of being used directly at inference.

Knowledge distillation transfers soft targets from the teacher to a smaller student~\cite{hinton2015distilling}, and learning with privileged information uses data, such as reports, available to the teacher during training but not during testing~\cite{vapnik2009lupi,lopezpaz2016unifying}. Existing CXR distillation methods transfer labels, embeddings, or explanations rather than spatial evidence. Termritthikun et al.~\cite{termritthikun2024explainable} distill a classifier with explainable outputs for on-device inference. X-Plain~\cite{sadhya2026xplain} distills the embeddings and text explanations of a vision--language model into an EfficientNet-B3 student. Thus, transferring spatial evidence raises a design question -- which regions should the teacher evaluate? Evaluating fixed proposals generated at warm-up is simple and cheap, but the student's evidence changes during training, so these proposals may no longer match the regions it currently emphasizes. DAgger~\cite{ross2011dagger} trains on states observed by the learner, on-policy distillation~\cite{agarwal2024onpolicy} supervises sequences generated by the student, and Soft Teacher~\cite{xu2021softteacher} weights student-proposed boxes by teacher scores in semi-supervised detection. It is unclear whether such learner-derived supervision helps pathology localization, where teacher queries are costly and teacher judgments of box extent are imperfect.

\begin{figure*}[t]
  \centering
  \includegraphics[width=0.88\textwidth]{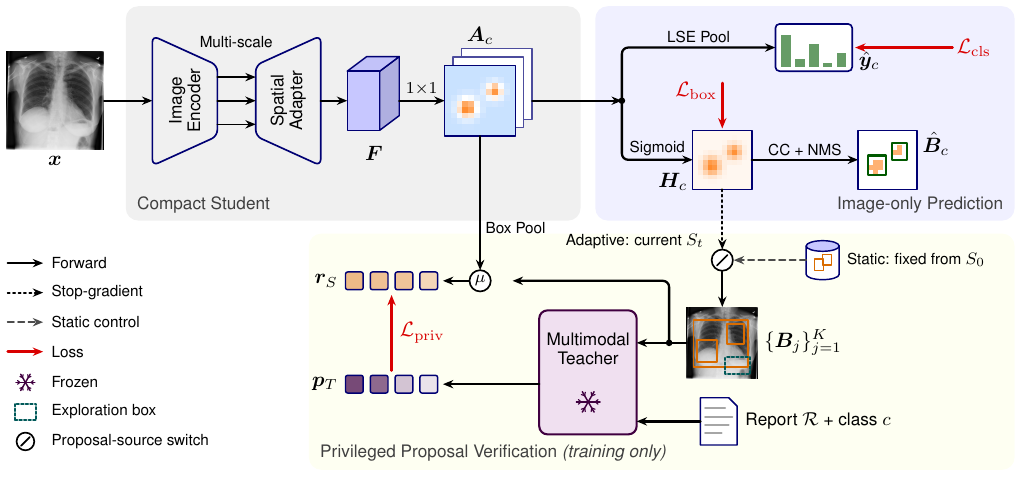}
  \caption{Overview of APED-CXR. The student predicts disease probabilities $\hat{\bm{y}}_c$ and boxes $\hat{\bm{B}}_c$ from evidence maps $\bm{A}_c$ using the image alone. In training, a frozen teacher verifies proposals $\{\bm{B}_j\}$; its support scores $\bm{p}_T$ supervise the box-pooled scores $\bm{r}_S$. The switch $\oslash$ selects proposals from the current student $S_t$ (APED-CXR) or the warm-up checkpoint $S_0$ (static comparator C2). Maps and boxes are schematic.}
  \label{fig:overview}
\end{figure*}

APED-CXR therefore uses a frozen multimodal teacher that reads the radiograph, report, and queried class to verify proposals from the current student, and turns its support scores into targets for the student's evidence maps (Fig.~\ref{fig:overview}). Since diagnosis and localization use the same maps, the student requires the image alone at test time, and the teacher's cost is restricted to training. As the teacher scores each proposal independently, its targets can amplify evidence in supported regions and dampen it when all proposals are poor. Our contributions are threefold: (1)~adaptive privileged verification, whose independent support targets directly modify the evidence maps used for image-only diagnosis and grounding; (2)~paired controls that separate adaptive feedback from query exposure, recency, scorer type, dense proposal coverage, and cached teacher-generated boxes, with the teacher cost of each; and (3)~a patient-disjoint held-out evaluation using box-level grounded F1 on PadChest-GR, with five paired seeds and patient-level bootstrap intervals, plus a secondary evaluation on NIH ChestX-ray14 boxes~\cite{wang2017chestxray8}.

\section{Method}
\label{sec:method}

\subsection{Evidence-Map Student}
\label{ssec:student}

An EfficientNet-B3 encoder~\cite{tan2019efficientnet} and a lightweight adapter compose features from strides 8, 16, and 32 of a radiograph $\bm{x}$ into a stride-8 map $\bm{F}$. A $1{\times}1$ convolution predicts the evidence logits and the bounded evidence map of class $c$,
\begin{equation}
  A_c(u,v) = \bm{w}_c^{\top}\bm{F}_{u,v} + b_c, \qquad H_c = \sigma(A_c),
  \label{eq:evidence}
\end{equation}
where $\sigma$ is the sigmoid function. Log-sum-exp (LSE) pooling~\cite{pinheiro2015image,wang2017chestxray8} over the $n$ valid, non-padded cells produces the disease probability,
\begin{equation}
  z_c = \frac{1}{\rho}\log\Big(\frac{1}{n}\sum_{u,v} e^{\rho A_c(u,v)}\Big), \qquad \hat{y}_c = \sigma(z_c),
  \label{eq:lse}
\end{equation}
where $\rho = 1$. Since there is no extra classifier, spatial supervision of $A_c$ also affects the diagnostic output. At inference, $H_c$ is thresholded, connected components are turned into boxes, and non-maximum suppression (NMS) with a box cap produces $\hat{\bm{B}}_c$. Boxes are generated only if $\hat{y}_c \geq \tau_c$, thus a single class might give multiple boxes or none.

\subsection{Proposals and Privileged Verification}
\label{ssec:verify}

For each explicitly positive image--class pair, proposals are extracted without backpropagation by applying the inference extraction at thresholds $\{0.2, 0.4, 0.6, 0.8\}$, merging duplicates, and keeping the top three components by mean evidence. One exploration box from a fixed multi-scale bank completes the set ($K=4$) and lets the teacher also support regions where the student has no focus yet. The student scores the proposal $B_j$ by its mean evidence, $r_S^{(j)} = \sigma\big(|B_j|^{-1}\sum_{(u,v)\in B_j} A_c(u,v)\big)$.
The frozen teacher, MedGemma 1.5 4B~\cite{sellergren2025medgemma}, was selected since it was trained on CXRs, works locally to satisfy the data agreements, and exposes answer likelihoods. It receives the radiograph $\bm{x}$ with $B_j$ drawn as an outline, the normalized box coordinates, the preprocessed report text $\mathcal{R}$, and the class name, and is asked whether the box is sufficiently localized to capture one visible occurrence of the finding. If $\ell_{\mathrm{yes}}$ and $\ell_{\mathrm{no}}$ are the log-likelihoods of two fixed answer tokens, the support score is
\begin{equation}
  p_T^{(j)} = \sigma\big((\ell_{\mathrm{yes}} - \ell_{\mathrm{no}})/T\big),
  \label{eq:pt}
\end{equation}
where the temperature $T$ is fitted on development data. The prompt is fixed before training since answer likelihoods depend on prompt format~\cite{zhao2021calibrate}. As scores are independent across proposals, a set of consistently bad proposals can be awarded uniformly low support.

\subsection{Training Objective}
\label{ssec:loss}

Let $m_{i,c}\in\{0,1\}$ be a mask that marks explicitly positive or negative labels in the report-derived labels; uncertain and unmentioned labels are masked rather than considered negative. The classification loss $\L_{\mathrm{cls}}$ is the average of the binary cross-entropy ($\BCE$) between $\hat{y}_{i,c}$ and $y_{i,c}$ over the labels with $m_{i,c}=1$.
On images with manually labeled boxes, the union $M_{i,c}$ of same-class boxes supervises $H_c$ through $\L_{\mathrm{box}}$, the sum of the $\BCE$ and Dice~\cite{milletari2016vnet} losses between $H_c$ and $M_{i,c}$; the filled boxes are coarse targets, but not lesion masks. The privileged loss is the average of the soft-target $\BCE$ first on the $K_{i,c}$ proposals of the pair and then on the set $\mathcal{P}$ of eligible pairs,
\begin{equation}
  \L_{\mathrm{priv}} = \frac{1}{|\mathcal{P}|}\sum_{(i,c)\in\mathcal{P}} \frac{1}{K_{i,c}} \sum_{j=1}^{K_{i,c}} \BCE\big(r_S^{(j)}, p_T^{(j)}\big),
  \label{eq:priv}
\end{equation}
and the total loss is $\L = \L_{\mathrm{cls}} + \beta\,\L_{\mathrm{box}} + \gamma\,\L_{\mathrm{priv}}$.

\subsection{Adaptive and Static Proposal Schedules}
\label{ssec:schedule}

Stage~A trains on diagnostic labels only; Stage~B additionally uses $\L_{\mathrm{box}}$ and produces a warm-up checkpoint $S_0$ for each seed. During Stage~C, each branch begins with the same $S_0$ and optimizer state. The static comparator extracts and scores proposals just once from $S_0$ but leaves them constant, while the current evidence inside them is pooled again at every update. The adaptive approach regenerates proposals from the current student $S_t$ every $N$ updates, scores the new proposals, and reuses cached scores for duplicate queries (the same image, class, report, and box). The branches share images, sampling order, loss weights, and update budget, so the only difference between them is the proposal source.

\section{Experimental Setup}
\label{sec:setup}

\subsection{Datasets}
\label{ssec:data}

We train on frontal adult PadChest radiographs~\cite{bustos2020padchest} with their preprocessed Spanish reports and report-derived labels (27\% annotated by physicians, the rest by a recurrent network). We employ the official stratified patient split of PadChest-GR~\cite{castro2025padchestgr}, which also covers the other PadChest studies, so no patient overlaps partitions. PadChest-GR provides boxes by up to two radiologists; its 3{,}185 training and 455 validation studies act as grounding-train and grounding-val, and its \PopStudies{} test studies form the primary held-out test set, never used for training or tuning. For a secondary cross-dataset set, we use the NIH ChestX-ray14 boxes~\cite{wang2017chestxray8}, where a radiologist boxed one queried finding per image; since the teacher was trained on ChestX-ray14~\cite{sellergren2025medgemma}, this set is not independent of the supervision. After auditing label mappings and per-class statistics, we retain only atelectasis (ATL), cardiomegaly (CMG), consolidation (CON), edema (EDM), pleural effusion (EFF), and pneumothorax (PTX); lung opacity and pneumonia are excluded because they overlap other classes. Distillation uses a fixed, patient-stratified subset of \DistillImages{} training images with at least one positive target finding; images of grounding-val and test patients are excluded before it is drawn.

\subsection{Controls}
\label{ssec:controls}

C0 uses diagnostic labels only, C1 extends them with manual boxes without a teacher, and C2 serves as the static comparator. C3 refreshes random boxes matched in number and size, so it receives as many teacher queries without learner-derived regions (query exposure). C5 scores a dense fixed bank of 32 boxes per pair once (coverage rather than refreshing), and C6 replaces the teacher with report-derived laterality and zone rules on current proposals (scorer type). C7 uses proposals from two refreshes earlier (recency), and C8 transforms boxes generated once by the teacher into targets on the fixed bank (generation rather than verification). C4 scores current proposals by their overlap with manual boxes (annotation-based feedback) and is thus assessed on grounding-train images only.

\subsection{Evaluation Protocol}
\label{ssec:eval}

The prespecified primary evaluation metric is class-aware box-level micro-gF1 at an intersection-over-union (IoU) threshold of 0.5 on the PadChest-GR test partition. Eligible pairs include \PopPos{} fully annotated positive pairs with \PopBoxes{} boxes from the designated primary reader and \PopNeg{} reliable-negative pairs; uncertain, unmentioned, and non-localizable positives are excluded. Boxes are accepted with $\hat{y}_c \geq \tau_c$ and matched one-to-one to same-class reference boxes by maximizing matches with IoU\,$\geq$\,0.5. Then micro-gF1 is defined as $2\,\mathrm{gTP}/(2\,\mathrm{gTP}+\mathrm{gFP}+\mathrm{gFN})$, where unmatched predictions, including all boxes on negative pairs, constitute gFP and unmatched reference boxes, gFN. We also report macro-gF1 and the diagnostic macro-AUC over known labels. NIH lacks reliable negatives and boxes only the queried finding, so we report class-conditional box recall at IoU\,$\geq$\,0.5 for ATL, CMG, EFF, and PTX (\NihBoxes{} boxes, \NihImages{} images, \NihPatients{} patients) using the ungated boxes of the annotated class. Results are the mean\,$\pm$\,standard deviation (SD) over five paired seeds for C1, C2, and Full and three for the other controls, and 95\% confidence intervals (CIs) for paired differences over their common seeds come from 2{,}000 patient-level bootstrap samples, conditional on the trained seeds and resampling patients jointly for both branches.

\subsection{Implementation Details}
\label{ssec:impl}

Images are padded to $512{\times}512$ pixels with the aspect ratio conserved, giving $64{\times}64$ evidence maps. We use AdamW with a weight decay of 0.01, a learning rate of $3{\times}10^{-4}$ during Stage~A and $10^{-4}$ afterwards, and cosine decay, with 32 diagnostic and 16 grounding-train images drawn per update. Stages A, B, and C run 40{,}000, 6{,}000, and 12{,}000 updates, with a refresh every $N=1{,}200$ updates ($R=\Refreshes$). We set $\beta=1$ and select $\gamma\in\{0.1, 0.3, 0.5, 1.0\}$, $\tau_c$, and the heatmap threshold per branch on grounding-val with the same budget; NMS uses an IoU of 0.5 and at most three boxes per class. The teacher runs in bfloat16 without gradients. All experiments use PyTorch on one NVIDIA A100 GPU (40\,GB), and latency is averaged over 100 timed images after warm-up.

\section{Results and Discussion}
\label{sec:results}

\begin{table}[t]
\centering
\caption{Teacher scoring of grounding-val proposals from image $\bm{x}$, report $\mathcal{R}$, class $c$, and box $B$; generated boxes are mapped to proposals by overlap. Acceptable: IoU\,$\geq$\,0.5 with a same-class box. AUROC: acceptable vs.\ other; Top-1: best-scored is acceptable; $\rho_{\mathrm{IoU}}$: Spearman correlation with IoU; All-bad: mean support if none is acceptable. Time: ms per image--class pair.}
\label{tab:teacher}
\small
\setlength{\tabcolsep}{1.7pt}
\begin{tabular}{@{}llccccr@{}}
\toprule
Mode & Inputs & AUROC\,$\uparrow$ & Top-1\,$\uparrow$ & $\rho_{\mathrm{IoU}}$\,$\uparrow$ & All-bad\,$\downarrow$ & ms \\
\midrule
Image only & $\bm{x}$,$c$,$B$ & 0.716 & 0.46 & 0.37 & 0.41 & 318 \\
Report only & $\mathcal{R}$,$c$,$B$ & 0.689 & 0.43 & 0.31 & 0.46 & 86 \\
Report rules & $\mathcal{R}$,$c$,$B$ & 0.664 & 0.40 & 0.28 & 0.38 & $<$1 \\
Generation & $\bm{x}$,$\mathcal{R}$,$c$ & 0.652 & 0.39 & 0.30 & \textbf{0.19} & 1{,}380 \\
Image+report & $\bm{x}$,$\mathcal{R}$,$c$,$B$ & \textbf{0.781} & \textbf{0.55} & \textbf{0.46} & 0.27 & 333 \\
\bottomrule
\end{tabular}
\end{table}

\begin{table*}[t]
\centering
\caption{Box-level grounding on the held-out PadChest-GR test partition (IoU\,$\geq$\,0.5; mean\,$\pm$\,SD over five seeds for C1, C2, and Full, three otherwise). $\Delta$: paired difference from C2 over common seeds (patient-bootstrap 95\% CI). All rows except C0 use the manual box loss.}
\label{tab:main}
\small
\setlength{\tabcolsep}{4pt}
\begin{tabular}{@{}llccccc@{}}
\toprule
ID & Spatial supervision & Micro-gF1\,$\uparrow$ & $\Delta$ vs.\ static [95\% CI] & Macro-gF1\,$\uparrow$ & AUC\,$\uparrow$ & Teacher GPU-h\,$\downarrow$ \\
\midrule
C0 & None (image-level labels only) & 0.140\,$\pm$\,0.014 & $-$0.165 [$-$0.188, $-$0.144] & 0.101 & 0.812 & 0 \\
C1 & Manual boxes only & 0.284\,$\pm$\,0.009 & $-$0.017 [$-$0.030, $-$0.004] & 0.201 & 0.809 & 0 \\
C2 & Teacher on fixed warm-up proposals ($S_0$) & 0.302\,$\pm$\,0.005 & -- & 0.215 & 0.813 & 5.9 \\
C3 & Teacher on refreshed random boxes & 0.311\,$\pm$\,0.022 & +0.006 [$-$0.010, +0.022] & 0.223 & 0.811 & 57.5 \\
C5 & Teacher on a fixed dense proposal bank & 0.308\,$\pm$\,0.008 & +0.004 [$-$0.016, +0.022] & 0.224 & 0.813 & 47.4 \\
C6 & Report rules on current proposals ($S_t$) & 0.311\,$\pm$\,0.009 & +0.006 [$-$0.011, +0.023] & 0.222 & 0.810 & 0 \\
C7 & Teacher on lagged proposals ($S_{t-2}$) & 0.322\,$\pm$\,0.022 & +0.018 [$-$0.001, +0.036] & 0.240 & 0.814 & 35.6 \\
C8 & Cached teacher-generated boxes & 0.310\,$\pm$\,0.008 & +0.005 [$-$0.012, +0.022] & 0.221 & 0.812 & 6.1 \\
\textbf{Full} & Teacher on current proposals ($S_t$) & \textbf{0.338}\,$\pm$\,0.018 & +0.036 [+0.022, +0.049] & \textbf{0.253} & \textbf{0.815} & 34.4 \\
\midrule
\multicolumn{7}{@{}l}{\textit{$^{\dagger}$Matched manual subset: distillation restricted to grounding-train images; $\Delta$ vs.\ C2$^{\dagger}$}} \\
C2$^{\dagger}$ & Teacher on fixed warm-up proposals ($S_0$) & 0.297\,$\pm$\,0.011 & -- & 0.213 & 0.810 & -- \\
C4$^{\dagger}$ & Manual-overlap score on current proposals & 0.313\,$\pm$\,0.009 & +0.016 [$-$0.000, +0.032] & 0.226 & 0.811 & -- \\
Full$^{\dagger}$ & Teacher on current proposals ($S_t$) & 0.306\,$\pm$\,0.005 & +0.009 [$-$0.007, +0.024] & 0.218 & 0.811 & -- \\
\bottomrule
\end{tabular}
\end{table*}

\subsection{Teacher Verification}
\label{ssec:res_teacher}

On grounding-val (Table~\ref{tab:teacher}), combined image and report verification gives the best proposal AUROC (\PilotBestAUROC) and top-1 hit rate (\PilotTopOne, candidate oracle recall \PilotOracle), and gives less support to all-bad sets than image-only or report-only scoring. Moving a proposal to an incorrect location with the report unchanged reduced its score by \MovedDrop{} on average, so the teacher responds to box location and not just report content. Report rules cover only \RuleCoverage\% of pairs, while generation, although conservative on all-bad sets, ranks proposals less accurately and returned a valid output for \GenValid\% of queries, so all teacher branches use combined verification.

\subsection{Adaptive versus Static Supervision}
\label{ssec:res_main}

In Table~\ref{tab:main}, manual boxes roughly double the micro-gF1 of C0, while static privileged supervision (C2) adds a further \StaticOverManual{} (95\% CI [\StaticManualLo, \StaticManualHi]). The adaptive procedure achieves \FullMicro, giving a paired gain of \GainStatic{} over C2 (95\% CI [\GainStaticLo, \GainStaticHi]) that is positive in all \NSeedsMain{} seeds (\GainStaticSeedMin{} to \GainStaticSeedMax). Diagnostic AUC remains between \AUCRangeLo{} and \AUCRangeHi{} across branches.

C3, C5, C6, and C8 stay close to C2, with paired differences of at most \CtrlNearStatic, and the adaptive procedure exceeds each by at least \FullOverCtrlMin{} (all 95\% CIs above zero). Refreshed random boxes (C3) receive as many teacher queries but are not derived from the student's evidence, so query exposure alone does not explain the gain. The dense bank (C5) is scored once and cannot react to the student's changing evidence, rules (C6) lack box extent, and generated boxes (C8) inherit the weaker ranking of generation (Table~\ref{tab:teacher}). Lagged proposals (C7) recapture part of the gain (\LagOverStatic{} over C2), but remain \LagGap{} below current proposals (95\% CI [\LagGapLo, \LagGapHi]), so recency contributes. In the matched subset, teacher-scored and annotation-scored proposals (C4) gain similarly over C1 (\SubsetFullGain{} versus \SubsetManualGain) and are not separated here (C4 $-$ Full$^{\dagger}$ \SubsetCfourVsFull, 95\% CI [\SubsetCfourVsFullLo, \SubsetCfourVsFullHi]), so the teacher is directionally consistent with annotation-based feedback, not shown to match it.

\begin{figure}[t]
  \centering
  \includegraphics[width=0.91\columnwidth]{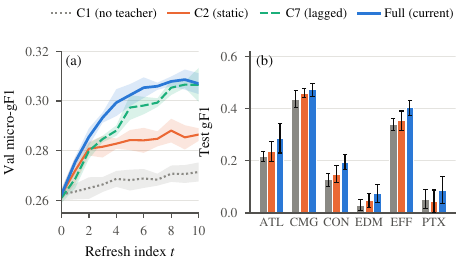}
  \caption{(a) Grounding-val micro-gF1 across refreshes and (b) per-class gF1 on the PadChest-GR test partition (mean\,$\pm$\,SD over seeds).}
  \label{fig:results}
\end{figure}

On grounding-val (Fig.~\ref{fig:results}a), C2 saturates after roughly two refresh intervals, while Full and C7 keep improving, with C7 about two refreshes behind. Relative to C1, the largest gains are for atelectasis, consolidation, and effusion (Fig.~\ref{fig:results}b), while cardiomegaly, where C1 already performs best, changes less; edema and pneumothorax have only \EDMPairs{} and \PTXPairs{} positive test pairs, so their estimates are unstable. The gain over C2 is comparable under a common threshold across branches (\CommonThrGain, 95\% CI [\CommonThrLo, \CommonThrHi]) and on the double-annotated subset averaged across readers (\DualReaderGain, 95\% CI [\DualReaderLo, \DualReaderHi]), and positive predictions without any box drop from \UngroundedStatic\% to \UngroundedFull\%. On NIH ChestX-ray14, adaptive supervision improves box recall from \NihStatic{} (C2) to \NihFull{} (\NihAdaptGain, 95\% CI [\NihAdaptGainLo, \NihAdaptGainHi]), while static supervision adds little over manual boxes (\NihStaticGain, 95\% CI [\NihStaticGainLo, \NihStaticGainHi]), matching its smaller in-domain gain.

\begin{table}[t]
\centering
\caption{Inference cost per image (A100, batch 1; ms, peak GB) and PadChest-GR test results; MedGemma generates boxes zero-shot from the image and class names.}
\label{tab:eff}
\small
\setlength{\tabcolsep}{2.2pt}
\begin{tabular}{@{}lcccccc@{}}
\toprule
Model & Params & GMACs & ms & GB & gF1\,$\uparrow$ & AUC\,$\uparrow$ \\
\midrule
MedGemma & 4B & -- & 1{,}910 & 9.6 & 0.189 & -- \\
APED-CXR & \textbf{11.1M} & \textbf{5.7} & \textbf{9.6} & \textbf{0.38} & \textbf{0.338} & \textbf{0.815} \\
\bottomrule
\end{tabular}
\end{table}

\subsection{Limitations and Efficiency}
\label{ssec:res_eff}

The main limitation is that the primary test set is from the training hospital. NIH ChestX-ray14 only partly addresses this, as it covers four classes without negatives and was used to train the teacher, so it is a secondary cross-dataset check rather than independent external validation. The supervision is also noisy: the teacher reads stemmed Spanish report text without stop words, and most PadChest labels were assigned automatically. Filled boxes are coarse targets, and an IoU\,$\geq$\,0.5 match neither measures lesion delineation nor establishes clinical faithfulness. Finally, the adaptive branch queries the teacher more often than C2, so they are compared at unequal teacher cost, although C3 receives even more queries and stays near C2.

The student needs \StudentParams M parameters and \StudentGMACs{} GMACs and takes \StudentLatency\,ms per image, including box extraction (Table~\ref{tab:eff}). Without the report and proposals, the zero-shot teacher grounds findings less accurately than the student and is about \GemmaSpeedRatio{} times slower. The cost of adaptivity appears during training: \GPUhFull{} teacher GPU-hours per seed, \TeacherCostRatio{} times that of C2, although \CacheHitFull\% of its queries are cache hits.

\section{Conclusion}
\label{sec:conclusion}

APED-CXR distills report-informed spatial evidence into a compact image-only CXR student using the teacher's verification of the student's own proposals. On held-out PadChest-GR data, refreshed learner-derived proposals outperformed fixed proposals and the other proposal sources on box-level grounding, the lagged ones recouped part of the gain, and the direction held on NIH ChestX-ray14. Future work will cut teacher cost by selective refreshing and use larger multi-institution test sets.

\balance
\section{Compliance with Ethical Standards}
\label{sec:ethics}

PadChest~\cite{bustos2020padchest} and PadChest-GR~\cite{castro2025padchestgr} were obtained from BIMCV under their research use agreements, and NIH ChestX-ray14~\cite{wang2017chestxray8} was downloaded from the NIH site \url{https://nihcc.app.box.com/v/ChestXray-NIHCC}. Ethical approval was not required for this retrospective study of de-identified data, in accordance with the data use agreements of the source datasets. The radiograph in Fig.~\ref{fig:overview} is a public NIH ChestX-ray14 image, and no PadChest images or reports are reproduced.

\section{Acknowledgments}
\label{sec:ack}

No funding was received for conducting this study. The authors have no relevant financial or non-financial interests to disclose. We acknowledge the NIH Clinical Center as the provider of the ChestX-ray14 data. ChatGPT (OpenAI) was used to assist with writing and Claude Opus~5.5 (Anthropic) to assist with coding; the authors reviewed and edited all AI-assisted output and take full responsibility for the content of this paper.

\bibliographystyle{IEEEbib}
\bibliography{references}

\end{document}